\documentclass[letterpaper, 10 pt, conference]{ieeeconf}
\IEEEoverridecommandlockouts
\usepackage{enumerate}
\usepackage{tabularx}
\usepackage{booktabs}
\usepackage{graphicx} 
\usepackage{amsmath}
\usepackage{amssymb}
\usepackage{mathtools}
\usepackage{algorithm}
\usepackage{algorithmic}
\usepackage{color}
\usepackage{hyperref}
\usepackage[nocompress]{cite}


\newcommand{\EE}{\mathbb{E}}

\title{Foundation Models for Rapid Autonomy Validation}
\author{
  Alec Farid$^*$\thanks{$^*$ Equal contribution.} \quad Peter Schleede$^*$ \quad Aaron Huang \quad Christoffer Heckman \\
  Zoox Inc. \\
  \texttt{\{afarid, pschleede, ahuang, checkman\}@zoox.com}
}

\begin{document}
\maketitle

\begin{abstract}
We are motivated by the problem of autonomous vehicle performance validation. A key challenge is that an autonomous vehicle requires testing in every kind of driving scenario it could encounter, including rare events, to provide a strong case for safety and show there is no edge-case pathological behavior. Autonomous vehicle companies rely on potentially millions of miles driven in realistic simulation to expose the driving stack to enough miles to estimate rates and severity of collisions. To address scalability and coverage, we propose the use of a behavior foundation model, specifically a masked autoencoder (MAE), trained to reconstruct driving scenarios. We leverage the foundation model in two complementary ways: we (i) use the learned embedding space to group qualitatively similar scenarios together and (ii) fine-tune the model to label scenario difficulty based on the likelihood of a collision upon simulation. We use the difficulty scoring as importance weighting for the groups of scenarios. The result is an approach which can more rapidly estimate the rates and severity of collisions by prioritizing hard scenarios while ensuring exposure to every kind of driving scenario.
\end{abstract}

\section{Introduction}
\label{sec:introduction}
As human operators are removed from autonomous vehicles and driving software matures to carry passengers, simulated behavior validation takes an ever more central role. Behavior validation entails justifying that the software stack responsible for executive control of the vehicle can achieve a desired performance target across the expected exposure of an operating design domain (ODD). Without a human operator either physically present or remotely assisting, the validation stack must both exercise all expected behaviors and return a set of concrete metrics in a trustworthy way.

As the size of the ODD increases, the necessary amount of validation increases, often at a rapid rate. For instance, unique road geometries, higher speed limits, quirks of driving norms at a new location, and agent behavior complexity all contribute to an ever-growing set of validations to perform. This growth is capable of scaling beyond the bounds of available resources if the set grows with all collected driving logs. An example of how quickly these costs can grow, especially for rare events, is discussed in Appendix~\ref{ap:cost_of_simulation_example}. High precision validation, especially when compared against human performance can be a significant expense and a challenge for development.

In this work, we focus on the setting where a developer has collected significant driving logs but seeks to prioritize their simulation. This prioritization may be used within a fixed simulation budget to maximize effective validation, to reduce a set of simulations in order to save compute spend or total validation latency, or a combination thereof. Prior works have performed searches to find adversarial conditions that stress the autonomy stack \cite{zhaoaccelerated, mining-1, mining-2, mining-3}. Our approach analyzes a fixed set of logs from real driving scenes, which provides insights into the system's performance across naturally occurring scenarios and their distributions.

\begin{figure}[t]
\begin{center}
\includegraphics[width=\columnwidth]{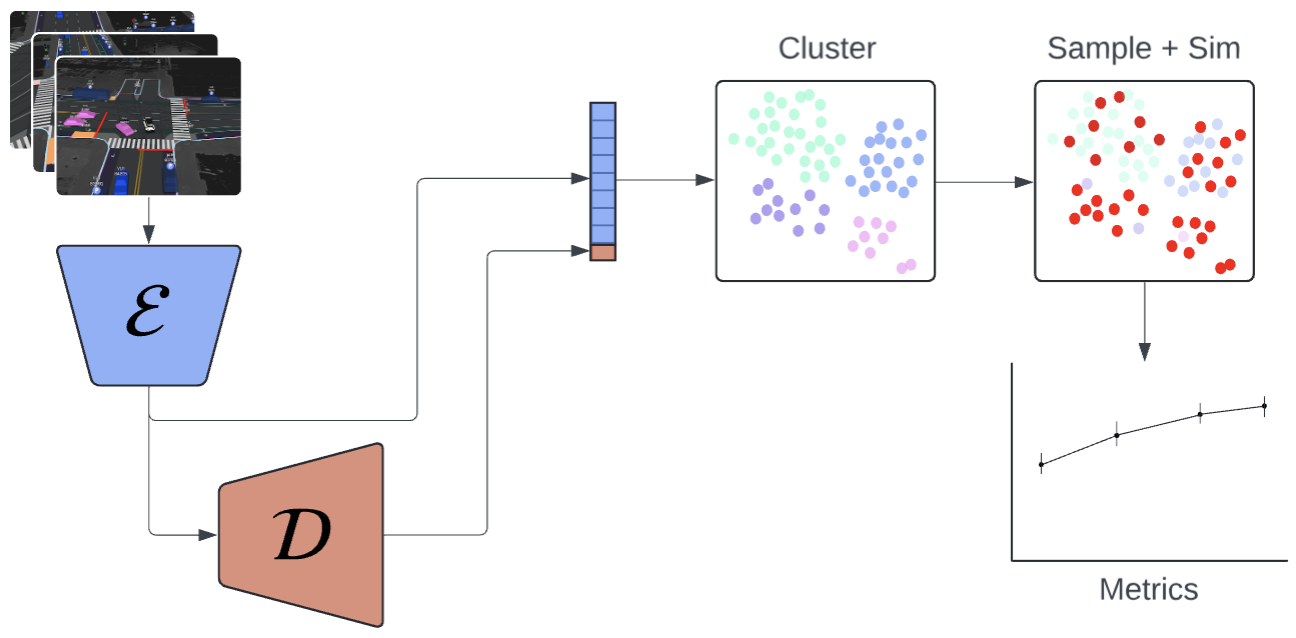}
\end{center} 
\vspace{-10pt}
\caption{Overview of the proposed validation process. We embed a scenario using a pretrained encoder $\mathcal{E}$. The embedding is concatenated with a difficulty score from a predictor head $\mathcal{D}$ which takes the embedding as input. Concatenated embeddings from a set of scenarios are clustered and weighted. Importance sampling and simulation is performed to compute validation metrics on autonomy (red dots are selected scenarios).
}
\label{fig:overview}
\vspace{-12pt}
\end{figure}

Core to answering the question ``is this software safe enough for operation without human intervention in this ODD?" are three criteria:
\begin{itemize}
    \item Simulation resources should be directed to diverse driving scenarios to cover the input distribution.
    \item Simulation resources should be directed to difficult driving scenarios to ensure they maximize the signal in the validation metric.
    \item The sampling method should return the exposure to each scenario type to calculate a weighted validation metric.
\end{itemize}

Self-supervised learning (SSL) allows for the learning of representations of input data that can be used for interacting with data and the relationships inherent in it. The pretraining process that is part of SSL also yields good starting points for fine-tuning powerful downstream models. In this paper, we describe a pretraining process used for driving data and two uses of this pretrained model for the behavior validation prioritization task. 
Our specific contributions are: (i) Training an SSL model for general driving behavior understanding (rather than only as a pretraining step as in \cite{forecastmae}). (ii) Validating an autonomy stack by combining complementary uses of the pretrained SSL model:  clustering based on scenario similarity and scoring scenarios based on their difficulty. (iii) Extensive evaluation of the validation technique.

\section{Related Work}
\label{sec:related work}

\textbf{AV Foundation Models.} In recent years, foundation models have undergone swift and remarkable advancements \cite{Firoozi2023FoundationMI, Zhou2023ACS, 10.1007/s10586-023-04203-7} spurred by the seminal Transformer paper \cite{vaswani2017attention} and the success of early models such as BERT, GPT, and T5 \cite{bert, liu2023summary, raffel2019exploring}. Foundation models are the dominant approach for NLP tasks \cite{Khurana2023}, and recent progress in multimodal foundation models \cite{xu2024surveyresourceefficientllmmultimodal, Cui_2024_WACV, zhang2024mmllmsrecentadvancesmultimodal} have shown impressive performance on modality understanding and generative tasks in the image \cite{liu2023visual, pmlr-v202-li23q, openflamingo, openai2021dalle, rombach2022stablediffusion, midjourney2022}, video \cite{videochat, sora, klingai2023, dreammachine2023, llama-vid, video-llava}, and audio \cite{qwen-audio, udio, stableaudio, audiocraft} domains. Given the notable success of MLLMs on these tasks, a reasonable extension is to apply multimodal foundation models to the robotics and AV domains which feature a variety of sensor modalities, embodied environments, and extensive sensor and actuation data \cite{Firoozi2023FoundationMI, Cui_2024_WACV}. Foundation models have been used to support language-conditioned imitation learning \cite{play-lmp, cliport, mcil, voltron}, reinforcement learning \cite{ada, palo-et-al}, value learning \cite{r3m, saycan, inner-monologue}, task planning \cite{nl2tl, chen-et-al}, and end-to-end control \cite{rt1, rt2, rt-x, pact, latte}. Modern embodied AI approaches materialize LLM-based agents in world simulators \cite{minedojo, saycan, code-as-policies, socratic-models} and instruct them to achieve goal states or maximize score \cite{voyager, ellm, vpt, embodiedgpt}. Specific to AVs, multimodal foundation models are useful for perception tasks \cite{driving-with-llms, drivegpt4, hilm-d, talk2bev} because of their powerful few-shot and in-context learning capabilities. There is also growing interest in utilizing their generative capabilities for photo-realistic simulation \cite{gaia-1, unisim}. On the planning and control front, approaches generally use combinations of modality encoders to project input data into aligned token representations that are fed to a reasoning backbone, which then produce output text and actions \cite{dilu, gpt-driver, drivegpt4, surreal-driver, rrr, languagempc, talk-to-the-vehicle}. Explainability is a notable benefit to many of these approaches, with many models capable of generating explanations of why they made a decision and the environmental factors to which they attend. 

\textbf{AV Validation \& Verification.} ISO 26262 is a functional safety standard for automotive electronic and electrical systems, focusing on lifecycle management and risk assessment. ISO 26262 sets the limit for the acceptable ratio of faults as 10 Failures in Time (FIT), meaning that given $10^9$ hours of operation no more than 10 faults should be observed. However, AVs are also generally unsafe to test on the road at that scale and modern AVs undergo extremely dynamic and fast-paced engineering cycles that necessitate frequent re-verification and validation \cite{rajabli, ma}. Simulation has emerged as a partial solution for AV V\&V that is rapid, scalable, and safe compared to real-world testing. Plenty of open-source simulators exist for AV testing \cite{torcs,sumo,gazebo,airsim,apollo} with some including benchmarks for autonomous driving quality \cite{carla_leaderboard, nuplan}. Despite being more scalable than real-world testing, simulated driving can still be prohibitively expensive at scale. A standard approach is to mine for difficult scenarios from existing driving logs \cite{zhaoaccelerated, mining-1, mining-2, mining-3}, or synthesize difficult or rare scenarios often using ML \cite{rajref17, maref74, maref71, maref80, maref75, maref87, okelly2018, bak2022} with \cite{diffusion_scenegen} showing off a controllable diffusion-based approach allowing users to specify desired scenario properties. Fault injection is another common approach for creating difficult test scenarios, with frameworks such as \cite{drivefi, tensorfi, binfi, asfault_0, asfault_1, mobatsim, koren2018} using algorithms like Bayesian Optimization to rapidly hunt for faults that lead to system failures. Other approaches specifically target ML-based components by inserting sensor noise \cite{rajabliref32, rajabliref38}, with ``white-box" frameworks like \cite{deepxplore, dlfuzz} adversarially mutating inputs in an attempt to maximize ``neuron coverage" in the system to be tested. Formal verification methods typically attempt to give rigorous guarantees on system safety by using mathematical tools like Lyapunov functions \cite{gangopadhyay_lyapunov, xu_ref_zhao, nguyen_fuzzy} and Satisfiability Modulo Theories \cite{smt_ref1, smt_ref2}, or logical modelling and proofs \cite{kamali, mehdipour, zhao_certification, liebenwen1, lienebwen2}.

\begin{figure}[t]
\vspace{3pt}
\begin{center}
\includegraphics[width=\columnwidth]{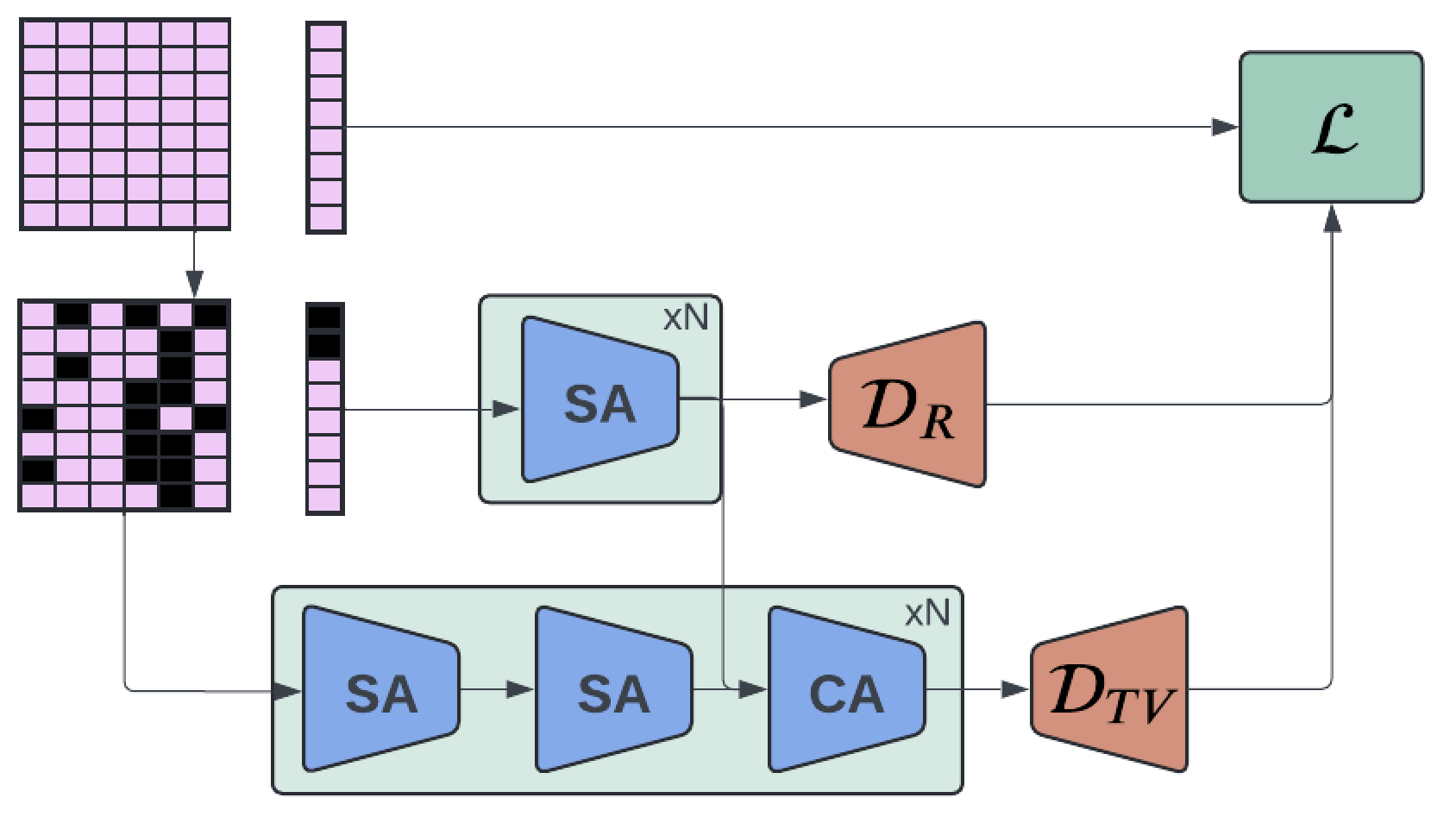}
\end{center} 
\vspace{-8pt}
\caption{The pretraining process. Time variant and time invariant inputs are randomly masked. They are then passed through the encoder, described in detail in \ref{subsec:encoder}. The embeddings are then passed through a decoder and a reconstruction loss is computed.}
\label{fig:pretraining}
\vspace{-10pt}
\end{figure}

\section{Pre-training}
\label{sec:pre-training}
We adopt the masked autoencoder (MAE) training objective. Portions of the scene input are masked, after which the partially masked inputs are encoded and decoded (see Figure~\ref{fig:pretraining} for an overview). We compute a reconstruction loss between the decoder outputs and the original, unmasked inputs. We also use a sparse representation of the driving scene rather than a birds-eye view rendering. This allows for high resolution representation of the inputs and a natural use of the transformer architecture we apply in the encoder. 
\subsection{Time Variant Inputs}
The number of time steps, $T$, is constant for each time variant input.

\subsubsection{Tracks}
The track tensor $X_T\in\mathbb{R}^{N_T\times T\times D_T}$ contains $N_T$ tracks from the perception system, including classes such as vehicles, pedestrians, cones, lane dividers, and others. For each track-time step, we encode the pose, velocity, acceleration, extents, and class into a single $D_T$-dimensional vector. We assign each unique track to a row in $X_T$. If there are fewer than $N_T$ tracks or a track is not visible at all time steps, $X_T$ is zero-padded where necessary.

\subsubsection{Traffic Signals}
The signal tensor $X_S\in\mathbb{R}^{N_S\times T\times D_S}$ contains $N_S$ signals where each time step has a $D_S$-vector encoding pose and a perception label for each signal. We use a map with signal locations, so no time steps are padded as they can be set with the pose and an UNKNOWN label.

\subsection{Time Invariant Inputs}
To encode the road network, we adopt an approach similar to \cite{forecastmae}. We extract vectorized polylines representing the centers of lanes and annotation marks such as stop lines, parking spaces, and crosswalks. The $N_Z$ polylines are represented with three tensors:
\begin{itemize}
    \item Coordinate frames for each polyline $X_F\in\mathbb{R}^{N_Z\times D_F}$ where each $D_F$-vector contains $x,y$ locations and sine and cosine of the rotation to the frame
    \item Class labels $X_L\in\mathbb{R}^{N_Z\times D_L}$ for one-hot encoded $D_L$-vector
    \item Sets of $S_Z$ polyline points $X_P\in\mathbb{R}^{N_Z\times S_Z\times D_P}$ expressed in the associated coordinate frame where each $D_P$-vector contains $x,y$ positions, width, and existence
\end{itemize}

\subsection{Masking}
As is typical in an MAE setup, we randomly sample inputs to mask with masking ratio $r$. Masking corresponds to replacing the encoded features' vectors with a 0-vector. For the time variant masks $M_T\in\mathbb{R}^{N_T\times T}$ and $M_S\in\mathbb{R}^{N_S\times T}$, we sample across the first two dimensions of the input independently. For the time invariant mask $M_Z\in\mathbb{R}^{N_Z}$, we randomly mask over the polylines. If a polyline is masked, the coordinate frame is left unmasked but the labels and points are all masked, including the existence. Note that at inference time, we do not mask the inputs.

\subsection{Model}
Our model is largely an encoder-style transformer as described in \cite{bert, forecastmae}. See Figure~\ref{fig:pretraining} for an overview of the architecture. We make some required modifications because: 
\begin{itemize}
    \item Text data has a clear ordering over the sequence while driving data has both temporal and spatial axes
    \item Vectorized road network polylines are easy to work with when collapsed to a single embedding per polyline \cite{vectornet}
    \item The inputs are not tokenized nor of varying sizes, so we project them into a shared space and back out to their original shape
    \item We let the road network embeddings aid in generation of time variant embeddings but not vice versa based on the assumption that road network geometries are independent of what is on them
\end{itemize}

\subsubsection{Encoder}
\label{subsec:encoder}
We begin by collapsing the road polyline points, $X_P$ into a single vector using a PointNet \cite{pointnet} as done in \cite{forecastmae}. They are then concatenated with the labels $X_L$ into a tensor $Y_R\in\mathbb{R}^{N_Z\times D_R}$. The three inputs, $X_T,X_S,Y_R$ and coordinate frames $X_F$ are each projected into a shared hidden dimension $D$ that is used throughout the encoder, referred to as $F_{proj}$. $F_{proj}$ acts as a positional embedding for $Y_R$. $X_T$ and $X_S$ are concatenated along the first dimension for a time variant input $Y_V\in\mathbb{R}^{(N_T+N_S)\times T\times D}$. The position embeddings for $Y_V$ are simple sin/cos encodings independently added over the first two axes, adding a notion of which time variant object a row describes and which time a column describes.

$Y_R$ is passed through a set of transformer encoder layers to obtain $Z_R$, the embeddings for all road polylines. $Y_V$ is passed through a factorized attention transformer similar to \cite{wayformer} where it self-attends over the spatial and time dimensions separately and is then cross attended with $Z_R$ at each layer. This yields $Z_V$, the embeddings for all object-time steps in $Y_V$.

\subsection{Loss}
We independently sample a mask with ratio $r_{\text{loss}}$ following the same sampling methodology as described previously in the masking section. For all inputs that are covered by this mask, we compute a reconstruction loss. Thus, the loss applies to both masked and unmasked inputs. The loss is a weighted sum with weights $\lambda$ over individual input types: 
\begin{equation*}
    \mathcal{L} = \lambda_T\mathcal{L}_T+\lambda_S\mathcal{L}_S+\lambda_R\mathcal{L}_R+\lambda_{\text{ego}}\mathcal{L}_{\text{ego}}.
\end{equation*}
In addition to the track loss $\mathcal{L}_T$, signal loss $\mathcal{L}_S$, and road network loss $\mathcal{L}_R$, we introduce a loss term $\mathcal{L}_{\text{ego}}$ for reconstructing ego, as tasks may require a strong representation of ego. Each loss consists of an L1 loss on $J$ continuous inputs $x_j$, and a cross-entropy loss on $M$ categorical inputs $x_m$:
\begin{equation*}
    \mathcal{L} = \sum^J_j \|\hat{x}_j-x_j\|_1 + \sum^M_{m}CE(\hat{x}_m, x_m).
\end{equation*}
Where $\hat{x}_j$ is a predicted continuous input and $\hat{x}_m$ is a predicted categorical input. 
For the models used in later sections, we set all $\lambda = 1$.
\section{Scenario Difficulty}
\label{sec:difficulty scoring}
As explored in \cite{He_2022_CVPR}, MAEs provide effective representations for transfer learning to supervised tasks. The MAE showed strong performance in both linear probing and full fine-tuning for image-based tasks, which influenced our approach. Considering our aim to utilize both the MAE latent space and a classification head output, we opted to keep the backbone frozen while training only the classification head.

\subsection{Data}
The data for the fine-tuning is similar to data used to train the difficulty model presented in \cite{bronstein23a}. We use a set of recorded driving logs and their results after simulation. For a given scenario, we apply a label of 1 if simulation resulted in a collision and 0 if it did not. In general, it is possible for a driving log to have a simulation result on many different software versions. If so, we add it to the dataset as multiple examples. Note that each would share identical inputs, as we encode the original recorded driving log. For scenarios that do not unimodally fail or succeed, the optimal score is then some real number between 0 and 1.

\begin{figure*}[ht]
\vspace{3pt}
\begin{center}
\includegraphics[width=2.0\columnwidth]{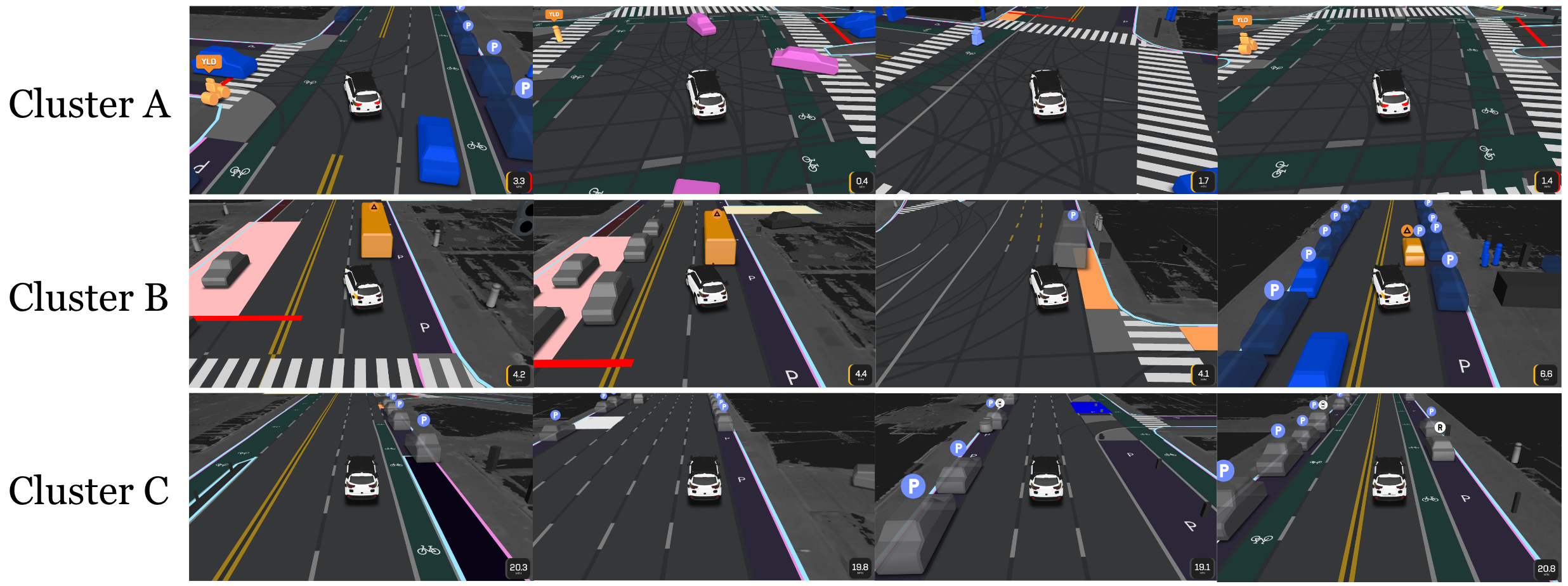}
\end{center} 
\vspace{-8pt}
\caption{Representative scenarios from representative clusters. Cluster A consists of scenarios where the ego vehicle is making an unprotected left turn and yielding to a pedestrian or bicyclist which is blocking the turn. Cluster B consists of double-parked vehicles that the ego vehicle is nudging around to the left. Cluster C consists of ego proceeding straight on a road with no moving vehicles around. \label{fig:cluster introspection}}
\vspace{-10pt}
\end{figure*}

\subsection{Fine-tuning}
We use a pre-trained encoder and perform a combination of pooling and concatenation over the outputs. We pool the ego, track, signal, and road network embeddings to create multiple aspects of the scene. These are then concatenated. An MLP is added converting from this combined embedding to a scalar output. We train with a binary cross-entropy loss.

While log divergence and other factors can lead to false positive collisions in our training set, the simulator is of sufficient quality where many simulated collisions are worth measuring.

\subsection{Inference}
At inference time, the model output can be considered a probability of collision as it is a value in $[0, 1]$. We let this represent a continuous notion of difficulty as we find that scenarios with a high likelihood of collision are, in some way, difficult for the driving stack. Empirically, after inspecting high and low difficulty predictions, we find that the predictions match human intuition and in some cases, flag examples already known to validation teams as challenging.

\section{Sampling Schemes}
\label{sec:sampling scheme}
In this section, we introduce DICE (Difficulty-based Importance sampling on Clustered Embeddings), a novel scenario sampling scheme. This approach leverages two key capabilities of the pre-trained foundation model. DICE increases the exposure to challenging and interesting scenarios without sacrificing the diversity of scenarios inherent in a large validation set. Note that all sampling schemes presented select scenarios without replacement in order to select the largest diversity of scenarios. 

\subsection{Clustering scenarios}
\label{subsec:clustering scenarios}
During pre-training, the model learns to represent the distribution of driving in the latent space. The result is that qualitatively similar scenarios tend to be close together in the latent space as shown in Figure \ref{fig:cluster introspection}. This motivates grouping scenarios within the latent space to form sets of self-similar clusters. The groups are not only self-similar, but represent qualitatively different kinds of behaviors i.e. there is some separation of behavior due to clusters.

During validation, it is crucial to ensure that every kind of scenario, especially those less common or more complex, is tested to validate changes in behavior comprehensively. After clustering scenarios, we can expose the autonomy stack to a representative sample from each cluster to cover each of the various behaviors on which validation is required. This intuition motivates a sampling technique where we uniformly sample scenarios across clusters (see Algorithm~\ref{alg:cluster subsampling}).

\begin{algorithm}
\caption{Uniform sampling across clusters \label{alg:cluster subsampling}}
\begin{algorithmic}[1]
\STATE $N$ scenario embeddings $\{z_i\}_{i=1}^N$, from the pre-trained backbone where $z_i$ has $D$ dimensions
\STATE $\{c_j\}_{j=1}^M \leftarrow \texttt{cluster}(\{z_i\}_{i=1}^N)$: cluster into $M$ groups
\STATE Sampled scenarios $S = \{\}$
\WHILE {below sampling budget}
    \STATE Sample $j$ from $\{1, \dots, M\}$
    \IF {$c_j$ is not empty} 
        \STATE Sample $k$ from $\{1,\dots, \texttt{size}(c_j)\}$ 
        \STATE Add scenario $k$ from $c_j$ to $S$
    \ENDIF
\ENDWHILE
\end{algorithmic}
\end{algorithm}

Testing does not need to cover every individual scenario within well-represented clusters, which are already abundant in the dataset. However, one of the primary challenges in validating autonomy stacks is that most driving scenarios are straightforward and do not significantly challenge the system. While it is still necessary to test these simpler scenarios, relying solely on them can lead to a slower validation. This challenge underscores the need for an additional signal to guide the testing process: scenario difficulty.

\subsection{Importance sampling}
\label{subsec:importance sampling}
We incorporate scenario difficulty (Section~\ref{sec:difficulty scoring}) by appending and weighting the scenario's difficulty score as part of the embedded space. Thus, clustering will separate based on scenario similarity and difficulty.

With clusters separated in terms of difficulty, we can score each cluster (e.g. according to an average, or even a particular percentile of the difficulty scores within the cluster). The score $w_i$ for cluster $c_i$ can then be treated as an importance weight upon sampling. The resulting sampling scheme, DICE, presented in Algorithm~\ref{alg:sampling scheme}, ensures that the autonomy stack is more frequently exposed to challenging scenarios, which are crucial for thorough validation.

\begin{algorithm}
\caption{DICE: Difficulty-based Importance sampling on Clustered Embeddings \label{alg:sampling scheme}}
\begin{algorithmic}[1]
\STATE $N$ scenario embeddings $\{z_i\}_{i=1}^N$, from the pre-trained backbone where $z_i$ has $D$ dimensions
\STATE $N$ scenario difficulty scores $\{d_i\}_{i=1}^N$, from the fine-tuned difficulty-scoring model
\STATE $\{\hat{c}_j\}_{j=1}^M \leftarrow \texttt{cluster}(\{\texttt{concat}(z_i, d_i)\}_{i=1}^N)$: cluster into $M$ groups
\STATE $w_i \leftarrow \texttt{score\_importance}(\hat{c}_i) \ \forall \ i \in \{1, \dots, N\}$: score each cluster 
\STATE Sampled scenarios $S = \{\}$
\WHILE {below sampling budget}
    \STATE Sample $j$ from $\{1, \dots, M\}$ according to probability distribution $\frac{\{w_i\}_{i=1}^M}{\sum_{j=1}^M w_j}$
    \IF {$c_j$ is not empty} 
        \STATE Sample $k$ from $\{1,\dots, \texttt{size}(c_j)\}$ 
        \STATE Add scenario $k$ from $\hat{c}_j$ to $S$
    \ENDIF
\ENDWHILE
\end{algorithmic}
\end{algorithm}

By adding a constant during $\texttt{score\_importance}$, we can limit the impact of the difficulty score on sampling. Consider the following definition for the importance weight:
\begin{equation*}
    \texttt{score\_importance}(\hat{c}_j) := K_0 + \texttt{mean}\big(\{d_k\}_{k=1}^{\texttt{size}(\hat{c}_j)}\big)
\end{equation*}
where $d_k \in [0, 1]$ is the difficulty of scenario $k$ in cluster $\hat{c}_j$ according to the fine-tuned difficulty-scoring model. For $K_0 \rightarrow \infty$, each cluster will have uniform weights. When $K_0 = 1$, the highest-scored clusters will have at most twice as much likelihood to be sampled as the lowest-scored clusters. Note that there is another natural sampling scheme relying exclusively on the difficulty scores. If we sample the most difficult scenario at each step, we will see the most difficult scenarios, see Algorithm~\ref{alg:difficulty sampling}. However, using this approach we cannot guarantee a good diversity of scenarios. 

\begin{algorithm}
\caption{Sampling highest difficulty scenarios \label{alg:difficulty sampling}}
\begin{algorithmic}[1]
\STATE $N$ sorted scenario difficulty scores $\{d_i\}_{i=1}^N$ from the fine-tuned difficulty-scoring model such that $d_1$ has the highest difficulty and $d_N$ has the lowest.
\STATE Sampling budget of $B$ scenarios 
\STATE Sampled scenarios $S = \{d_i\}_{i=1}^B$
\end{algorithmic}
\end{algorithm}

\section{Experiments}
\label{sec:experiments}
This section details our training datasets and procedure. Once trained, we have a shared backbone model which produces a latent space for all scenario inputs. We then use the classification head to generate a scenario difficulty score and pool a portion of the embedding to produce the latent space used for scenario similarity. We validate our sampling scheme and compare it to random sampling. 

\subsection{Training details}
\label{subsec:training details}
We train the foundation model on 14 million driving snippets which are 10s each. The loss weights reconstruction evenly across 4 categories: (i) ego's state, (ii) agents' states, (iii) traffic signal states, and (iv) road features' states, see Section~\ref{sec:pre-training} for further details on the inputs and reconstruction loss. The model used in this experimental section has 34m parameters with a 64-dimensional latent space. The model took around 13 hours to train on an AWS p5.48xlarge node (8 GPUs each with 80GB of HMB3 GPU memory). We performed 1 pass over the training set with a batch size of 416 (this batch size was picked to maximize GPU utilization). The learning rate was set to $3e$-$4$ and we used the schedule-free optimizer \cite{defazio2024road}. 

We then freeze this pre-trained backbone and add an MLP classification head to the un-pooled embedding. Note the combined backbone and classification head have a single output, see Section~\ref{sec:difficulty scoring} for further details on the fine-tuning process. We train the classification head with 80,000 driving snippets using a uniform-weight cross-entropy loss. The data has a 50/50 split of collisions and not collisions corresponding to difficulty-1 and difficulty-0 scenarios respectively. The data is disjoint from the test set used to generate the results in this section. Note that freezing the pre-trained backbone allows us to significantly speed up inference on the model in order perform the sampling scheme. In particular, it only requires one inference pass through the backbone to get both the embedding space used to cluster as well as the input to the classification head. 

\begin{figure}[t]
\vspace{3pt}
\begin{center}
\includegraphics[width=1.0\columnwidth]{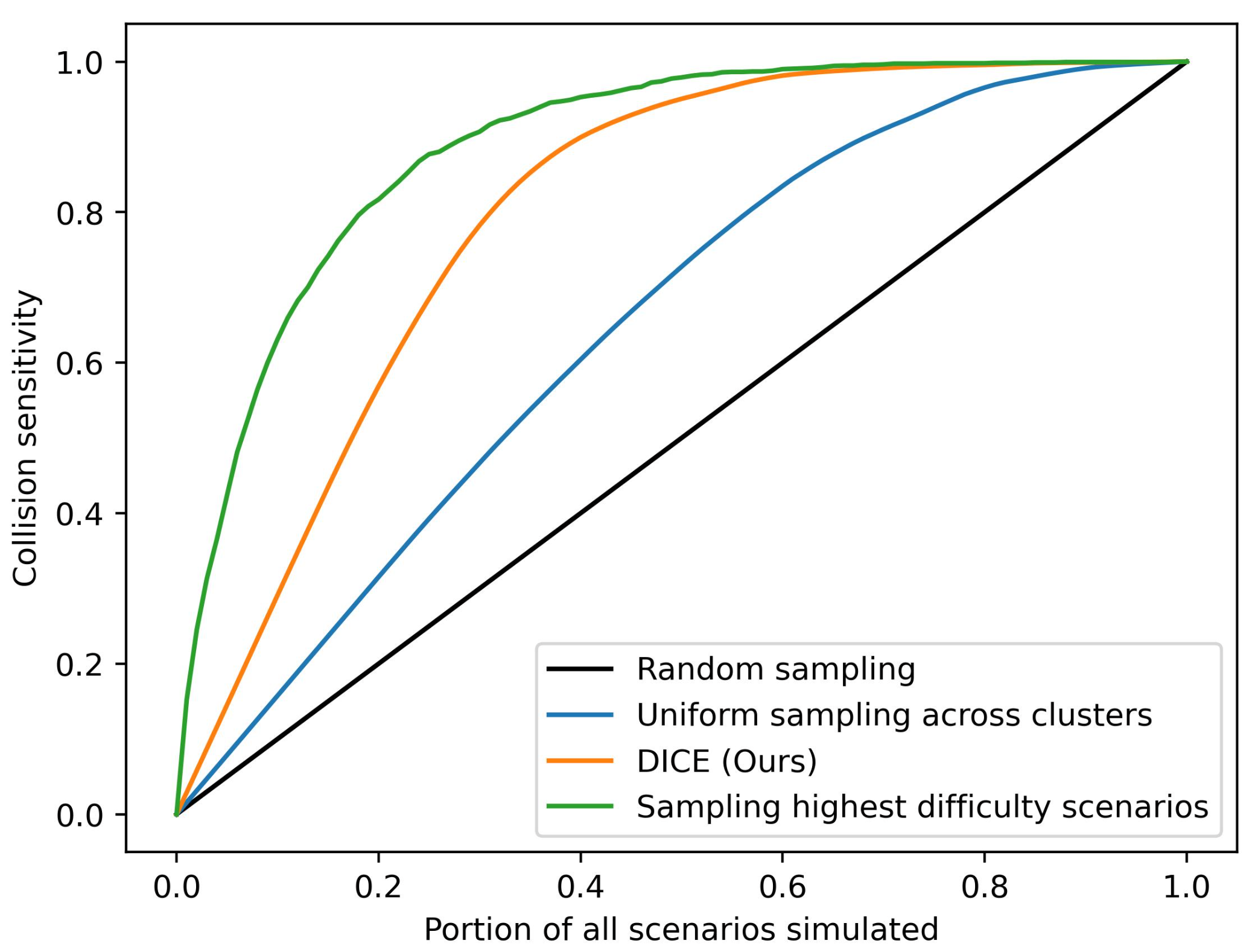}
\end{center} 
\vspace{-8pt}
\caption{Comparison of the proposed sampling schemes' expected proportion of collisions found. We compare random sampling with uniform sampling of clusters (Algorithm~\ref{alg:cluster subsampling}), DICE (Algorithm~\ref{alg:sampling scheme}), and pure difficulty-based sampling (Algorithm~\ref{alg:difficulty sampling}). \label{fig:sensitivity}}
\vspace{-10pt}
\end{figure}

For the embedding space used to cluster groups of similar scenarios, we mean-pool the ego state embedding over the full time-interval of the input scenario. The result is a 64-dimension embedding corresponding to ego's state throughout the input scenario. Note that even though other portions of the embedding are not included, ego's reconstruction is closely tied to nearby agents and road features and thus important features of the input scenario are also well represented in this embedding space. 

\subsection{Autonomy validation performance}
\label{subsec:autonomy validation performance}
Following DICE (Algorithm~\ref{alg:sampling scheme}), we produce a latent space for each scenario by concatenating the backbone embedding and difficulty score. The difficulty-scoring model was never trained on any data used in this section; this is to emulate a validation process where a new version of an autonomy stack is tested in simulation for the first time and we must select high-yield scenarios. On the a new version of the autonomy software, we can use the existing scenarios to re-train or fine-tune the difficulty-scoring model. 
The full scenario set used for validation ($N_\texttt{full}$) is 800,000 driving scenarios which corresponds to almost 100 days of continuous driving and approximately 50,000 miles driven. We validate our sampling scheme using the ground truth where we re-simulate every scenario. Upon re-simulation of all driving scenarios, we found 1300 simulated collisions ($C_\texttt{full}$). Suppose this large set of simulations is enough to provide high confidence in the desired metrics. In particular, the goal is to estimate the rate of simulated collisions with high precision so we may determine if changes to the autonomy stack improve or worsen the rate of simulated collisions. 

We first compare the sampling scheme against random sampling of scenarios from the full validation set. Upon random sampling of $N_\texttt{random}$ scenarios, the number of simulated collisions we should expect to see is given by
\begin{equation*}
 \EE [C_\texttt{random}] = C_\texttt{full} \frac{N_\texttt{random}}{N_\texttt{full}}.
\end{equation*}
Figure~\ref{fig:sensitivity} shows a comparison of the sampling schemes compared with random down-sampling. Note that as we approach sampling the entire simulation set, we approach $C_\texttt{full}$ collisions found because sampling is performed without replacement. We plot the portion of simulated collisions found (i.e. the sensitivity) as a function of the portion of the full dataset sampled. We also compare against a sampling strategy based solely on the difficulty scores. 

We see strongest performance, in terms of finding collisions, when only using difficulty scores to sampling scenarios. However, when we compare the distribution of samples selected from Algorithm~\ref{alg:cluster subsampling}, we find groups of scenarios are completely missed by pure difficulty-based sampling. In particular, Algorithm~\ref{alg:difficulty sampling} fails to select scenarios in $8\%$ of the clusters from Algorithm~\ref{alg:cluster subsampling}. Each cluster is well represented when running DICE (Algorithm~\ref{alg:sampling scheme}). Another challenge with using Algorithm~\ref{alg:difficulty sampling} is that it does not elicit a clear way to estimate the total rate of collisions in the full dataset without relying on historical data. Both Algorithm~\ref{alg:cluster subsampling} and \ref{alg:sampling scheme} have a natural way to estimate this statistic by assuming the sample from each cluster is representative of that cluster and scaling the number of events in that cluster by the inverse of the portion of that cluster which was sampled. 

The goal of the approach we present is to balance sampling from challenging scenarios with over-representation of common driving scenarios. With more collisions from diverse scenarios, we can gain a better understanding of the performance of the autonomy stack. However, it is still vital to validate the autonomy stack on all kinds of driving scenarios in order to ensure there is not a pathological behavior in an edge case.  

\section{Conclusions and Limitations}
\label{sec:conclusion}

The method described here demonstrates the value of self-supervised pre-training for scenario representations and provides an introspectable sampling approach that achieves meaningful improvements in simulation efficiency and provides multiple axes toward achieving further improvements. Several limitations exist in the proposed approach. Using prior collisions for training data ties the difficulty scoring to the simulation and autonomy stack. While we can mitigate this by using a rolling window of data or online learning, future work could investigate other methods of labeling difficulty that are independent from the autonomy stack. Additionally, the reliance on a scenario set that covers the distribution of driving from which to sample creates challenges for academic or smaller companies in generating sufficient data. Future work could study synthetic generation of realistic data to create this set or estimating likelihoods of data as a basis for exposure calculations. The clustering itself can also benefit from research into training embeddings for clustering, rather than using the embeddings that fall out of MAE pre-training.

\appendix
\subsection{Cost of simulation example}
\label{ap:cost_of_simulation_example}

To illustrate how expensive validation can be, consider a hypothetical example validating that a driving stack violates a safety metric (measured in miles per violation) no more than once per $N$ miles using a real-time simulator. If all roads have a 35 mph speed limit and we wish to simulate 10 times the desired metric for confidence, this requires $10N/35 \approx 0.286N$ driving hours of simulation. At time of writing, the on-demand cost for an AWS g5.4xlarge node is $\$1.624$ per hour \cite{aws_ec2_g5} and we assume one node fits one concurrent simulation. This yields a cost of $\$0.464N$. For a violation that should occur fewer than once every $50{,}000$ miles, this would cost roughly $\$23{,}200$ for each code change. 

There are many nuances to effective large scale simulation that are not included here, such as validating separately in different regions of a map, running multiple simulations per node, faster than real time simulation, node startup and shutdown costs, data transmission costs, limitations on pipeline width, and simulation fidelity.

\section*{ACKNOWLEDGMENTS}
\noindent We thank Vincent Spinella-Mamo and Trevor Herrinton for helpful feedback on the paper and Nathan Shemonski, Andrew Crego, and Seth Aaron for useful discussions on the sampling approaches.
\bibliographystyle{IEEEtran}
\bibliography{main.bbl}

\end{document}